\documentclass[letterpaper, 10 pt, journal, twoside]{ieeetran}

\IEEEoverridecommandlockouts                              % This command is only needed if 
\usepackage{graphics} % for pdf, bitmapped graphics files
\usepackage{graphicx}
\usepackage{amsmath} % assumes amsmath package installed
\usepackage{amssymb}  % assumes amsmath package installed
\usepackage{algorithm}
\usepackage{eucal}
\usepackage[usenames,dvipsnames]{xcolor}
\usepackage{algpseudocode}
\usepackage{verbatim}
\usepackage{tabularx}
\usepackage{mathrsfs}
\makeatletter
\let\NAT@parse\undefined
\makeatother
\usepackage[
colorlinks,
linkcolor=black,
citecolor=black,
filecolor=red,
urlcolor=blue]{hyperref}
\usepackage{balance}
\usepackage{booktabs} % for toprule/bottomrule
\usepackage{tabularx} % some extra functions for tables such as stretching on width

\algrenewcommand\algorithmicindent{0.5em}

\usepackage[numbers,sort&compress]{natbib}
\newcommand{\etal}{\textit{et al.}~}

\begin{document}
\title{
Multi-Robot Multi-Room Exploration with Geometric Cue Extraction and \\Circular Decomposition
}

\author{Seungchan Kim$^{1}$, Micah Corah$^{1}$, John Keller$^{1}$, Graeme Best$^{2}$, Sebastian Scherer$^{1}$ 
\thanks{Manuscript received: July, 25, 2023; Revised November, 2, 2023; Accepted November, 27, 2023.}
\thanks{This paper was recommended for publication by Editor M. Ani Hsieh upon evaluation of the Associate Editor and Reviewers' comments. This work is supported by Defense Science and Technology Agency Singapore.}
\thanks{$^{1}$ S. Kim, M. Corah, J. Keller, and S. Scherer are with Robotics Institute, Carnegie Mellon University, Pittsburgh, PA 15213, USA.  \tt{\{seungch2, micahc, jkeller2, basti\}\linebreak[0]{}@andrew.cmu.edu}}
\thanks{$^{2}$ G. Best is with School of Mechanical and Mechatroninc Engineering,  University of Technology Sydney, Ultimo NSW 2007, Austrailia. \tt{graeme.best@uts.edu.au}}
\thanks{$^{a}$ Video: \href{https://youtu.be/zUtK1hh2Tpo}{https://youtu.be/zUtK1hh2Tpo}}
\thanks{Digital Object Identifier (DOI): see top of this page.}
}

\algrenewcommand\algorithmicrequire{\textbf{Input:}}
\algrenewcommand\algorithmicensure{\textbf{Output:}}

\maketitle

\markboth{IEEE Robotics and Automation Letters. Preprint Version. Accepted Nov, 2023}{Kim \MakeLowercase{\textit{et al.}}: Multi-Robot Multi-Room Exploration with Geometric Cue Extraction and Circular Decomposition}

%%%%%%%%%%%%%%%%%%%%%%%%%%%%%%%%%%%%%%%%%%%%%%%%%%%%%%%%%%%%%%%%%%%%%%%%%%%%%%%%
\begin{abstract}
This work proposes an autonomous multi-robot exploration pipeline that coordinates the behaviors of robots in an indoor environment composed of multiple rooms. Contrary to simple frontier-based exploration approaches,  we aim to enable robots to methodically explore and observe an unknown set of rooms in a structured building, keeping track of which rooms are already explored and sharing this information among robots to coordinate their behaviors in a distributed manner. To this end, we propose (1) a geometric cue extraction method that processes 3D point cloud data and detects the locations of potential cues such as doors and rooms, (2) a circular decomposition for free spaces used for target assignment. Using these two components, our pipeline effectively assigns tasks among robots, and enables a methodical exploration of rooms. We evaluate the performance of our pipeline using a team of up to 3 aerial robots, and show that our method outperforms the baseline by $33.4\%$ in simulation and $26.4\%$ in real-world experiments. \textcolor{blue}{ \href{https://youtu.be/zUtK1hh2Tpo}{[Video]$^a$}}
\end{abstract}

%%%%%%%%%%%%%%%%%%%%%%%%%%%%%%%%%%%%%%%%%%%%%%%%%%%%%%%%%%%%%%%%%%%%%%%%%%%%%%%%
\section{Introduction}

\IEEEPARstart{M}{ulti-robot exploration} \cite{Burgard2005, braga2017} in unfamiliar, unknown environments has attracted attention in the robotics research community, due to its potential to accomplish duties faster than a single robot, and its wide applicability in tasks including search \& rescue operations \cite{search-and-rescue}, hazardous source detection in turbulent environments \cite{hazardous}, and planetary missions \cite{schuster2019towards}. Recently, in light of the DARPA Subterranean Challenge, there is a growing attention on exploration of large underground and indoor environments by teams of robots, with realistic communication constraints, sensor coverage, and compute conditions \cite{kulkarni2022subt, Scherer:2022, agha2021nebula}.
In this work, we aim to develop a more structured multi-robot autonomous exploration pipeline for operation in indoor environments, taking advantage of geometric properties of structures in a building.

Specifically, we aim to build algorithms for multiple robots exploring inside a building composed of multiple rooms, whose locations and sizes are not known in advance. Rather than following a frontier-based approach~\citep{yamauchi1998frontier}, we explicitly model the geometry of rooms to enable a methodical exploration of structural environments, empowering robots to explore the rooms one by one, observing each room in-turn before moving on to another. Furthermore, we also want coordinated behaviors such that robots select rooms to avoid redundant observations by multiple robots.%\textcolor{teal}{, and we are interested in performing detailed observation of these confined spaces (rooms) with a primary sensor, such as an RGB camera, at close range~\cite{best2022rss}.}

Why do we want this type of multi-room exploration? First, rooms are usually the parts of the building where meaningful target objects are placed in, compared to corridors and hallways.
Thus, they warrant focused attention during exploration. Second, rooms are non-overlapping and structural units that constitute the building; the entire indoor space can be segmented using a room-oriented search. One thing to consider is that, it is inefficient to assign multiple robots to the same small room; thus, we need a coordinated target assignment scheme for multi-robot, multi-room exploration.

\begin{figure}[!t]
    \centering
\includegraphics[width=0.95\linewidth, height=0.55\linewidth]{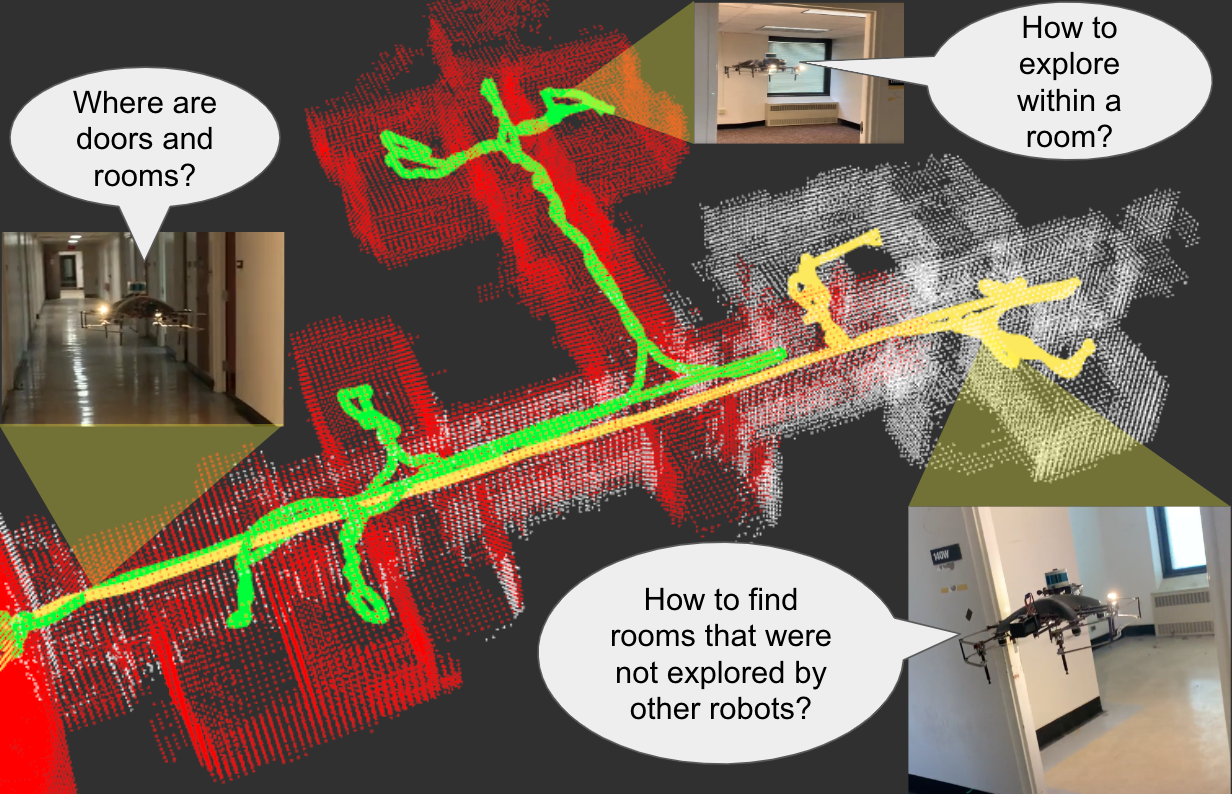}    \caption{We present an autonomous multi-robot exploration pipeline that coordinates the behaviors of robots exploring multiple rooms in a building. Using our pipeline, robots explore different rooms in a methodical manner. The traveled trajectory and area covered by each robot is visualized with different color.}
    \label{intro-figure}
\end{figure}

To this end, we propose our method, Multi-Robot Multi-Room (MRMR). Unlike learning-based geometry prediction for exploration, our method does not require an offline dataset of representative environments to learn from. Instead, we use a simple yet effective and generalizable technique to extract geometric cues that indicate the potential locations of doors and rooms, only from 3D point cloud data. Our pipeline processes data observed by LiDAR sensors onboard, converting 3D point clouds into a 2D binary occupancy grid map, and then into a 2D distance transform map. Using the distance transform map, robots perform real-time geometric analysis to discover structural cues that signal the doors and rooms. Robots update global plans and execute local planners accordingly, actively searching for unreached doors and unexplored rooms.

We also propose the idea of representing free spaces within rooms with circles. We show that circular decomposition is a compact representation of the environment for robots exploring rooms and sharing necessary information with other robots via communication. 

We evaluate our autonomous exploration pipeline using multiple unmanned aerial vehicles (UAV), both in simulation and real-world experiments. Built upon the multi-robot exploration and planning component of the complete autonomy stack \cite{best2022rss} of Team Explorer, which showed most successful exploration by aerial robots in the final round of DARPA SubT Challenge competition, our methods record significant performance gain in fast discovery and exploration of multiple rooms, and coordination of behaviors for multiple robots.

In summary, our contributions are:
\begin{itemize}
    \item An autonomous multi-robot exploration pipeline that extends prior works in \cite{best2022rss} to coordinate behaviors of robots in a building composed of multiple rooms.
    \item Incorporation of two new modules for multi-robot multi-room exploration: (1) a geometric cue extraction method that detects the locations of doors and rooms from 3D LiDAR point cloud data, and (2) circular decomposition of spaces for room representation, target assignments, and communication.
    \item Empirical validation of our multi-robot exploration pipeline via simulated and real-world experiments.
\end{itemize}

\section{Related Work}

\subsection{Multi-robot Exploration}
Multi-robot exploration problems have been studied with various approaches. Many prior works  \cite{Burgard2005, yamauchi1998frontier} view this problem as an assignment of frontiers (the boundaries between known and unknown space), where robots explore environments by continuously moving toward nearby, unexplored frontiers. Other approaches include sampling-based \cite{sample-based-motion-planning}, information-theoretic \cite{charrow2015information, submodular-micah2019}, graph search-based \cite{dang2020graph}, recursive tree-based search \cite{plume2019}, and sub-map merging \cite{MUI-TARE}. Recent works include hybrid approaches, such as combining frontier-based approach and graph-search \cite{best2022rss}. 

Multi-robot exploration research can also be categorized by whether decision-making is centralized or decentralized. In centralized schemes \cite{luna2011efficient, gul2022centralized}, a central entity plans out tasks for a team of robots with an access to the global information of the environment, which could hypothetically produce globally optimal solutions.
However, a single failure of a robot or communication link could lead to the failure of the entire system. Instead, we follow the decentralized \cite{li2020graph, zhai2021decentralized} schemes, which are more robust to the single point of failure. Each robot makes decisions and optimizes trajectories based on its own understanding of the environment, with realistic communication constraints among robots.

\subsection{Space Partitioning and Decomposition}
In robotic exploration research, space partitioning approaches decompose space into subparts or partitions and seek to cover the whole space by consecutively exploring the partitions. \citet{voronoi-based-space-partitioning} proposes Voronoi-based partitioning to coordinate multi-robot exploration, dividing the entire space into a set of polygons. \citet{unsup-clustering-space} propose using unsupervised clustering for multi-robot coordination. \citet{Voronoi-deep-rl} uses the combination of deep reinforcement learning and Voronoi-based partitions to improve coordination strategies for multi-robot exploration. 

In this work, we focus on decomposing the free space into a set of circles. Previously, Gao \etal \cite{Gao19} proposed generating circles from raw point clouds for safe online trajectory optimization in cluttered environments. Ren \etal \cite{bubble-planner} improves this idea by generating corridors with larger sphere volumes and receding schemes that enable high-speed trajectory planning. Most recently, Musil \etal \cite{SphereMap} demonstrates incrementally-built segmented graph of spheres enable safe and flexible flights. \cite{SphereMap} also shows the compressed version of the graph structure enables efficient communication in multi-robot setting. While the previous works used the circular or spherical decomposition in the context of safe trajectory planning, we focus on generating circular representations in the context of indoor room exploration, coupled with geometric cue extraction from distance transform map; we also focus on the multi-robot exploration, showing circular representation is an effective information scheme to be shared for decentralized behavior coordination of multiple robots.

\begin{figure*}[!t]
    \centering
\includegraphics[width=0.9\textwidth, height=0.32\textwidth]{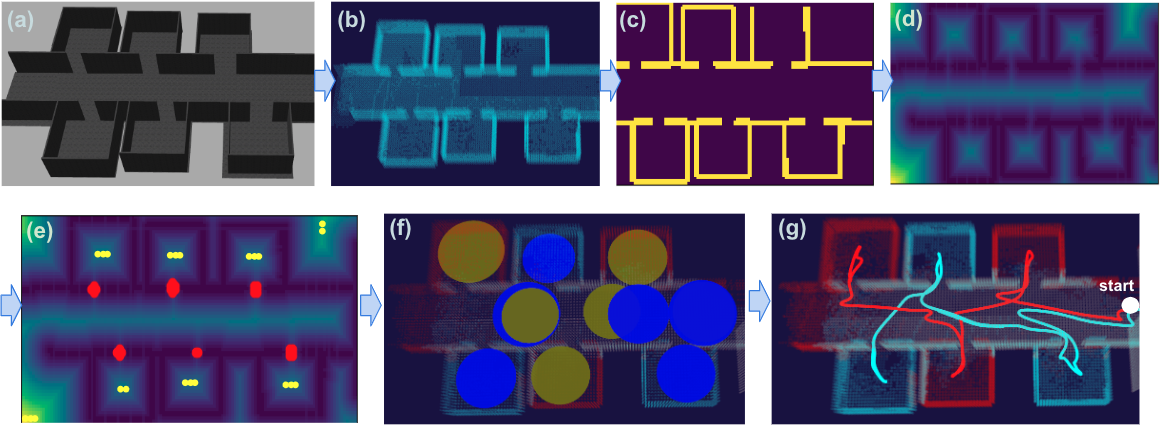}
    \caption{Overview of our exploration pipeline. (a): An indoor environment composed of multiple rooms. (b): The robot generates 3D occupancy voxel grid map from 3D point cloud data. (c): It flattens 3D voxel gid map into 2D binary map, and applies median filtering. (d): The robot generates 2D distance transform map (e): It computes saddle points (red) and local maxima (yellow) from the distance transform map. (f): Free space is decomposed into circular representations. In multi-robot setting, the robot shares the circles it explored (blue), and the circles that other robots have explored (yellow circles). (g): Trajectories traveled by two robots are visualized with different colors, which are results of the coordinated behaviors via shared circular representations.}
    \label{fig:diagram}
\end{figure*}

\subsection{Room Detection}
Division of a floorplan into rooms, or identifying the potential location and size of rooms is essential to room-based search and exploration. Most popular approach to detect rooms is Voronoi-based approach by \citet{thrun1998}, which utilizes distance transform and Voronoi graph to find critical points in the map to detect passages to the room. \citet{wurm2008coordinated} uses this idea to divide the space map into segments that can correspond to individual rooms, and generate Voronoi graph to assign targets for each robot in the multi-robot teams, in a centralized manner. 

Other works include graph-based partitioning which uses topological map to cluster nodes and build higher-level hierarchical map \cite{brunskill2007topological}, sparse graph-based approach to build 3D skeleton diagrams using 2D Voronoi Diagram \cite{sparse-3d-topologicalg-graphs}, or feature-based room segmentation \cite{sjoo2012semantic} which learns features and geometric shapes that match with the characteristic of rooms. More recently, scene graph-based methods were proposed: \citet{S-graphs+} proposed factor-graph with hierarchical layers and room segmentation scheme using 3D LiDAR point clouds; \citet{3d-dynamic-scene-graphs} proposed 3D dynamics scene graph, with multiple layers in a hierarchy representing different levels of semantic structures including rooms, followed by \citet{Hydra} which features improved hierarchical scene graph generated real-time. %where rooms are composed using lower-level layer graph nodes like walls, floors, and ceiling. 

In this work, we focus on LiDAR processing approach that enables high-speed, low-compute processing onboard and robust exploration in light-degraded environments, as opposed to learning-based approaches that often require a large amount of computations and pretraining on offline datasets. We revisit the idea of distance transform \cite{thrun1998}, but improve this idea by using a different cue detection method and space decompositions that are compatible with state-of-the-art fully autonomous multi-robot exploration algorithms. %in unknown, unfamiliar, degraded, communication-limited environments. 

\section{Problem Definition}
\label{problem-def}

% Robots move through environment
Consider a team of $n$ robots ($i=1,2,...,n$) that are exploring an indoor environment. Denote the trajectory traveled by each robot as $\xi_i$. The robots build a map of the environment with LiDAR, which is represented with a 3D occupancy voxel grid in a shared coordinate system (each voxel represents a cube with side length $0.2$m).
% Sensor model
The robots observe surfaces of the environment with a primary sensor, such as a camera or RGBD sensor with limited field of view. The intersection between the occupied voxels of the LiDAR map and the primary sensor field of view is denoted by the voxel grid $O_i$ (for robot $i$). The observation model for the primary sensor can be defined based on the specifications of the robot and the application; we model the primary sensor as a forward-facing fish-eye camera with a 5m detection range, and $170^{\circ}$ field of view, modelling occlusions with ray casting, as in~\cite{best2022rss}.

Let us assume that there are total $K$ rooms to be explored in the building, and denote $V^\mathrm{rm}_j$ ($j=1,2,...,K$) as voxels within each room.
Then, the total number of voxels in these rooms that are observed by the primary sensor of robot $i$ is
\begin{align}
\Bigl|\underset{j=1,...,K} {\bigcup}{(O_i \cap V^\mathrm{rm}_j)}\Bigr|
\label{room-voxel-objective}
\end{align}
($|\cdot|$ is the number of voxels).
In this work, we will reward robots for exploring rooms in a building rather than increasing total coverage by traversing corridors and hallways.

In the multi-robot setting, our objective is to maximize the union of such voxels, explored by all robots collectively.
Each robot finds its own trajectory $\xi_i$ seeking to maximize the union of observed voxels in rooms by all robots,
\begin{align}
\xi_1^*, \xi_2^*, ..., \xi_n^* = \underset{\xi_1, \xi_2, ..., \xi_n}{\arg \max} \Bigl|\underset{i=1..n}{\bigcup} \bigl(\underset{j=1..K} {\bigcup}(O_i \cap V^\mathrm{rm}_j)\bigr)\Bigr|
\label{eq:multi-robot-objective}
\end{align}
given a fixed time. The robots will solve \eqref{eq:multi-robot-objective} approximately and in a decentralized manner, so that each robot will plan its own path $\xi_i^*$ while communicating with other robots.

\section{Distributed Exploration Method}
%\gb{Fig.~\ref{fig:diagram} needs to be referenced somewhere, probably at the start of this section}
In this section, we present our autonomous distributed exploration pipeline, which coordinates the behaviors of multiple robots exploring in a building composed of multiple rooms. The pipeline overview is displayed in Fig.~\ref{fig:diagram}. We first explain preliminary background on the autonomous robot exploration baseline \cite{best2022rss}, which we build upon and improve (Sec.~\ref{sec:baseline}). Then, we describe the two main building blocks of our method, geometric cue extraction (Sec.~\ref{sec:geometric_cues}, Alg.~\ref{alg:extractCues}) and circular decomposition of free space (Sec.~\ref{sec:spherical_decomposition}, Alg.~\ref{alg:Spheres}). We also explain the multi-robot communication and target assignment (Sec.~\ref{sec:communication}, Alg.~\ref{alg:target}), and finally explain how they all come together to form our Multi-Robot Multi-Room (MRMR) method (Sec.~\ref{sec:exploration_system}, Alg.~\ref{alg:MRMR}). 

\subsection{Preliminary: Autonomous Exploration Baseline}
\label{sec:baseline}
The baseline method we use is the open source autonomous aerial robot exploration pipeline \cite{best2022rss} developed by Team Explorer, to compete in DARPA Subterranean Challenge. This baseline enables robots to navigate in a wide range of challenging underground or indoor environments such as a mine, subway, tunnel, or cave, with limited communication, sensor coverage, and light conditions. 

The baseline method leverages LiDAR sensors to discover the geometry of surrounding environments, using OpenVDB \cite{open-vdb} as a data structure for representing map occupancy grids, and SLAM solutions generated by Super Odometry \cite{super-odometry}. As the robot moves through the environment, it estimates which voxels have been visually observed according to the camera model in Sec III. with an OpenVDB map.

The path planning component of the baseline method, which we focus on in this work, can be loosely categorized as a frontier-based exploration approach with graph search and selection heuristics.
%\micah{"using vision and range sensors" The sensors don't make the viewpoints.}
Using vision and range sensors, the robot generates a set of \textit{viewpoints} at the frontiers. The viewpoints are scored with heuristics, and the viewpoints with high scores are selected. Then, paths are planned to reach the viewpoints. RRT-Connect \cite{RRT-connect} is used to find a feasible global path to viewpoints, and A* graph search and motion primitives are used for local path planning.

For multi-robot exploration, onboard communication hardware and DDS networking are used. The plans for robots are coordinated implicitly by sharing knowledge of the world. This includes knowing the take-off locations of each robot, and communicated shared map in a global reference frame.

\subsection{Extracting Geometric Cues of Doors and Rooms}
\label{sec:geometric_cues}

The first component of our method is detection of doors and rooms via geometric cue extraction, as shown in Alg.~\ref{alg:extractCues}. The intuition for this algorithm is that \emph{saddle points} on the distance transform are approximately equivalent to the locations of doors. We also extract local maxima and their distances to closest wall to decompose free spaces. 

\begin{figure}[!t]
    \centering
\includegraphics[width=0.95\linewidth]{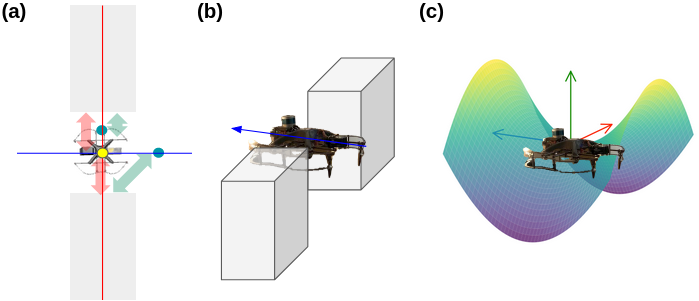}    \caption{(a) Saddle point (yellow) is farthest from occupied cells compared to other points along the wall axis (red), while closest to the occupied cells compared to other points along the perpendicular axis (blue) (b) Side-view of drone at a door (c) Drone at saddle point in distance transform space.}
    \label{saddle-points}
\end{figure}

The robot incrementally obtains 3D point cloud observations from onboard sensors which it uses to maintain a voxel grid map. The 3D voxel grid map $O$ is flattened into 2D binary map $B$ (occupied cells: $1$, unknown/free cells: $0$) by setting the point in 2D as occupied if any voxel within the range of height $z \in [z_{\text{low}},z_{\text{high}}]$ is occupied. In practice, we set $z_{\text{low}}=0$ and $z_{\text{high}}=1.8$. The robot applies median filtering\footnote{Median filtering of kernel size [1,3] and [3,1]} to the binary map $B$, and then converts $B$ into 2D distance transform map $M$, which computes the minimal distance to closest occupied cell for all pixels. Then the robot obtains a second-order partial derivative matrix of the distance map, or hessian $H$, from which it can compute saddle points $P^\mathrm{sadd}$ and local maxima $P^\mathrm{max}$, using the determinant\footnote{We empirically found that the condition $\det(x,y)<-0.1$ (instead of $\det(x,y)<0$) works well in practice to find saddle points. Similarly, we set $f_{\mathrm{xx}}(x,y)<-0.1$ as a threshold for detecting local maxima.}: 
%\micah{I adjusted typesetting to use upright letters with mathrm and such.}
\begin{align}
\det(x,y) = f_{\mathrm{xx}}(x,y) f_{\mathrm{yy}}(x,y) - (f_{\mathrm{xy}}(x,y))^2, 
\label{determinant}
\end{align}
and with $P^\mathrm{sadd}$ and $P^\mathrm{max}$ defined as
\begin{align}
\begin{cases}
(x,y) \in P^\mathrm{sadd} & \hspace{-3mm}\text{if}\ \det(x,y) < -0.1 \\
(x,y) \in P^\mathrm{max} & \hspace{-3mm} \text{if}\ \det(x,y) >0\ \& \ f_{\mathrm{xx}}(x,y) < -0.1
\end{cases}
\label{saddle-localmax}
\end{align}

\begin{algorithm}[t]
\caption{ExtractCues}\label{alg:extractCues}
\begin{algorithmic}[1]
\Require Voxel Grid Map Data $O$
%\State Create 2D binary map $B$ by taking $z \in [z_{\text{low}},z_{\text{high}}]$
\State 2D binary map $B$ $\leftarrow$ Take $O$ where $z \in [z_{\text{low}},z_{\text{high}}]$
\State Distance map $M$ $\leftarrow$ \textit{distanceTransform(filter$(B)$)}
\State Hessian matrix $H \gets \textit{Hessian}(M)$
\State Obtain two lists $P^\mathrm{sadd}, P^\mathrm{max} \gets $ from $H$ using Eqn.4
%\State Obtain a set of saddle points (doors) $P^\mathrm{sadd}$ from $H$
%\State Obtain a set of local maxima (center) $P^\mathrm{max}$ from $H$
\State $\Delta=[\;]. \hspace{2mm} \forall c \in P^\mathrm{max}$, distance $\delta \gets $ $M(c), \Delta.\text{append}(\delta)$

%\State For each $c\in P^\mathrm{max}$, obtain distance $\delta$ between $c$ to the closest wall using distance map $M$. This set is $\Delta$
 \Ensure $P^\mathrm{sadd}$, $(P^\mathrm{max},\Delta)$  
\end{algorithmic}
\end{algorithm}

A local maximum $m \in P^\mathrm{max}$ is a point, whose distance $\delta \in \Delta$ to the wall is the highest compared to other neighboring pixels. Pairs of $(m,\delta)$ will be used as centers and radii of circles, in the free space decomposition. 

The more interesting part is the use of saddle points (Fig.~\ref{saddle-points}). A saddle point refers to a critical point that is not a local extremum; any point that is a relative minimum along one axis and relative maximum along another perpendicular axis is classified as saddle points. In our setting, a door (yellow dot in Fig.~\ref{saddle-points}(a)) can be classified as a saddle point\footnote{In practice, multiple saddle points can be detected for a single door; however, points within distance $\epsilon_d=1.0m$ are handled the same.}, because it is farthest from occupied cells(walls) compared to other points placed along the wall-axis (red), while it is closest to the occupied cells(walls) compared to other free points placed along the perpendicular axis (blue). %Lastly, there are some saddle points that are not doors; for example, a narrow passageways between walls that are not doors can be detected as false positives, but they are negligible in practice. 
%\micah{Do we classify all saddle points as doors? How do we deal with saddle points that are not doors?}

\subsection{Circular Decomposition of Free Space}
\label{sec:spherical_decomposition}

\begin{figure}[!t]
    \centering
\includegraphics[width=1.0\linewidth]{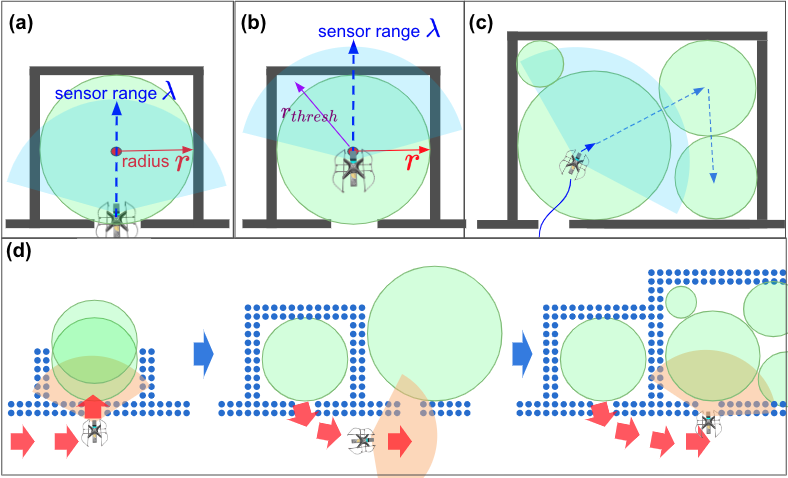}\caption{(a) We represent a room using a circle for robot observation. (b) We let the radius of the circle, $r$, is smaller than the sensor range $\lambda$ for effective observations. (c) A larger room is represented by multiple circles. (d) The robot updates, merges, splits the circles with new sensor information.}
    \label{sensor-coverage-circles}
\end{figure}

We represent the free space in rooms by decomposing it into circles. The motivation behind this is that visiting the centers of the circles will enable the robot to observe most of the room. Due to the limited camera FoV, some parts may be missed. However, in practice, the robot usually turns when going to and from the centers. If full visual coverage is particularly important, a spinning behavior can be added at each center, but we found this unnecessary in the considered scenarios. We note $360^{\circ}$ FoV can also help for full coverage. 

%\gb{begin by motivating why we need to decompose world into circles -- the idea being that visiting the centre of the circles will enable the robot to observe most of the room. due to the limited camera FOV, some parts may be missed. however in practice the robot will usually turn when going to and from the centres. if full visual coverage is particularly important, then a spinning behaviour could be added at each centres, but we found this unnecessary in the considered scenarios.} \gb{as noted in rebuttal -- 360 deg FOV would also help} \gb{see wording in rebuttal}

%We represent the \textcolor{red}{free} space by decomposing it into \textcolor{red}{circles}, where each circle's center and radius is a pair of local maximum $m \in P^{\max}$ and distance $\delta \in \Delta$ to the closest wall. Note that $(P^{\max},\Delta)$ is the output of Alg.~\ref{alg:extractCues}. 
%We argue that \textcolor{red}{circles} are compact and effective representation for room exploration. 

As in Fig.\ref{sensor-coverage-circles}(a)(b), we can generate a circle that is tangent to the inner wall of the room, and make the robot reach the center of the circle to observe the surroundings with a camera of sensor range $\lambda$; during this process, it is important to maintain that radius $r$ is smaller than a threshold value. For example, if $r>\lambda$, even if the robot goes to the center, there must exist a part in circle that is not covered by the sensor even within the FoV. In practice, we set the radius threshold for circle-split $r_{thresh}=2.5m$ ($r \leq r_{thresh}$), which is admissible with our sensor range $\lambda=5.0m$. In a larger room like Fig.\ref{sensor-coverage-circles}(c), the Alg.\ref{alg:extractCues} often outputs a longer distance $\delta > \lambda$. In this case, we split the circle into a set of circles with smaller radius $r$, and let the robot reach the centers of smaller circles one by one.

\begin{algorithm}[t]
\caption{UpdateCircles}\label{alg:Spheres}
\begin{algorithmic}[1]
\Require $C, (P^\mathrm{max}, \Delta)$ %New spheres $Q$
\State \textbf{for} {each $(m, \delta)$ pair:} \Comment{$m\in P^\mathrm{max}, \delta \in \Delta$}
\State \hspace{1mm}$c \gets GenerateCircle(m,\delta)$
%\State generate a sphere $s$ of center $c$ and radius $\delta$\\
\State \hspace{1mm}\textbf{for} {each $c' \in C$:}
\State \hspace{1.3mm} $r_1\gets\delta, r_2\gets c'.r, d\gets dist(c.center, c'.center)$
%\State $r1$ and $r2 \gets $ radius of $s$ and $s'$
%\State $d \gets$ distance between centers of $s$ and $s'$
\State \hspace{2.2mm}\textbf{if} $d<0.5(r_1+r_2)$:
$C \leftarrow C \cup \{c\} \text{ if } r1>r2$ %\textit{\Comment{if spheres are too close}}
%\State \hspace{2mm} choose / save the larger one between $s$ and $s'$
\State \hspace{0.8mm} \textbf{elif} $d<\epsilon(r_1+r_2)$: $c''\gets Merge(c,c'); C \leftarrow C \cup \{c''\}$%\textit{\Comment{if medium-distanced}}
%\State \hspace{2mm} merge the spheres by weighted-average them
\State \hspace{0mm} \textbf{if not} $c.merged:$ $C \leftarrow C \cup \{c\}$
%\State \textbf{if} $s$ isn't merged to any pre-existing $s' \in S$:
%\State \hspace{2mm} simply add $s$ to $S$
\State $\forall c\_ \in C, SplitCircle(c\_) \text{ if } c\_.r > r_{thresh}$
%\For{each new incoming sphere $q \in Q$}
%\For{each pre-existing sphere $s \in S$}
%\State $(s_x,s_y),(q_x,q_y)$: $(x,y)$ positions of center of $s,q$
%\State $(s_x,s_y) \gets$ $(x,y)$ coordinates of center of $s$
%\State $(q_x,q_y) \gets$ $(x,y)$ coordinates of center of $q$
%\State $r_s, r_q$: radius of $s$ and $q$
%\State $d \gets$ distance between $(s_x,s_y)$ and $(q_x,q_y)$
%\State \textbf{if} $d<0.5(r_s+r_q)$:
%\State \hspace{4mm}$s\leftarrow q \text{ \textbf{if} } r_s < r_q$
%\State \hspace{3mm} \textbf{if} $r_1 < r_2$:
%\State \hspace{6mm} replace $s$ with $q$
%\State \textbf{else if} $0.5(r_s+r_q)<d<(r_s+r_q)$:
%\State \hspace{3mm} 
%$x' = (r_ss_x+r_qq_x)/(r_s+r_q)$
%\State \hspace{3mm}
%$y' = (r_ss_y+r_qq_y)/(r_s+r_q)$
%\State \hspace{3mm}
%$r'=((r_s+r_q+d)/2) \cdot (d/(r_s+r_q))$
%$x=\frac{(r_1p_x+r_2q_x)}{(r_1+r_2)}, y=\frac{(r_1p_y+r_2q_y)}{(r_1+r_2)}$
%\State\hspace{3mm} $r=\frac{r_1+r_2+d}{2} \cdot\frac{d}{r_1+r_2}$
%\State \hspace{3mm} $s$ $\gets$ sphere with center $(x',y',1)$, radius $r'$
%\EndFor
%\State \textbf{if} $q$ isn't merged to any $s$ of the $S$
%\State \hspace{3mm} add $q$ to $S$
%\EndFor
\Ensure $C$
\end{algorithmic}
\end{algorithm}

The circular decomposition of the space is not static, as the robot updates the map by incrementally obtaining more information of the surrounding (e.g. discovers a new wall). As shown in Fig.~\ref{sensor-coverage-circles}(d), the robot updates the size of the circles so that they adjoin the newly observed walls, merge the circles if they overlap significantly, and split the circles if their radii are above the threshold $r_{thresh}$. This process of maintaining and updating a set of circles is shown in the Alg.\ref{alg:Spheres}. Given each pair of local maxima and distance, the robot generates a circle $c$ of center $m$ and radius $r_1=\delta$, which is compared with the circles $c'$ of center $c_t$ and radius $r_2$ in the previous set of circles $C$. If they overlap too closely, we save the larger one between the two; if distance $d$ between centers of $c$ and $c'$ are smaller than $\epsilon(r_1+r_2)$ where $\epsilon=0.95$, we merge these two into a new circle $c''$.\footnote{Although the merging methods vary, we empirically find that setting new center as $\frac{(r_1 \cdot m + r_2 \cdot c_t)}{(r_1+r_2)}$ and radius as $\frac{(r_1+r_2+d)\cdot d}{2(r_1+r_2)}$ effective.} We simply save $c$ if this isn't merged to any of the previous circles in $C$. Lastly, we split the circles if the circle's radius is above the threshold $r_{thresh}=2.5$, by setting the first split circle's radius as $2.5$ and the other circle to be adjoining to the first one.

% When the robot moves straight along the corridor, the geometry beyond the wall is unknown, so it assumes that there is a large open space, as in Fig.~\ref{spheres}(c). When observations are made through the door into the room, a partial map of the room is formed. As in Fig.~\ref{spheres}(d), the robot updates the sphere so that it adjoins the walls. In Fig.~\ref{spheres}(e), the robot has finished exploring the first room, and starts collecting data on the right side of the first room, adjusting the size and location of a new sphere accordingly. 

% While rooms are represented by spheres, they are not equivalent with rooms; in fact, any free space between walls or surrounded by walls can be represented by spheres. For instance, corridors are also filled with spheres. In practice, their existence doesn't affect the algorithm, as the target sphere is chosen after a door is reached. (See Sec.~\ref{sec:exploration_system})

\subsection{Multi-Robot Communication and Target Assignment}
\label{sec:communication}

As discussed in Sec.~\ref{sec:geometric_cues}, the robot extracts geometric cues and detects saddle points in 2D as potential doors, and it also represents rooms with circular decomposition of spaces as discussed in Sec.~\ref{sec:spherical_decomposition}. We send this 2D saddle points and centers of circle to 3D, by simply setting\footnote{Any $z$ around the middle of the range $(z_{low}, z_{high})$ is fine.} their $z$ value as 1.0, and update the set of doors $D$ and circles $C$. For distributed target assignment among multiple robots, we utilize the doors and circles as information to be shared. Each robot not only maintains $D$ and $C$, but also updates a set of doors $D_\mathrm{r}$ and circles $C_\mathrm{r}$ it has reached; as shown in Communication() function of Alg.~\ref{alg:target}, $C_\mathrm{r}, D_\mathrm{r} $ are shared with other robots bidirectionally. Receiving $C_\mathrm{r}, D_\mathrm{r}$ from other robots, the robot updates a set of doors and circles, $C_\mathrm{o}, D_\mathrm{o}$, that were reached by other robots. The information shared for communication, $D_\mathrm{r}$ and $C_\mathrm{r}$, are just a set of points in 3D coordinates and a set of pairs of point and radius. This compact representation enables efficient communication among robots.  

%the Sets of doors (lists of points in 3D coordinates) and sets of spheres (lists of points and radii) serve as a compact representation of the explored space. 

%\footnote{Communication for this approach is quadratic in the number of robots. Our approach seeks to use compact representations and messages to enable coordination across small teams of robots.} 
% \begin{figure}[!t]
%     \centering
% \includegraphics[width=0.9\linewidth]{Figures/target-spheres.png}\caption{Example image of spherical decomposition of space for distributed target assignments. The spheres that were reached by the first and second robot are colored blue and yellow, respectively.}
%     \label{target-spheres}
% \end{figure}

\begin{algorithm}[t]
\caption{TargetDoor, TargetCircle, Communication}\label{alg:target}
\begin{algorithmic}[1]
\Require $C$, $D$, $C_\mathrm{r}$, $D_\mathrm{r}$, $C_\mathrm{o}$, $D_\mathrm{o}$,  $\epsilon_d=1.0, \epsilon_c=1.5$
%\begin{itemize}
%\item a set of spheres it already reached, $S_\mathrm{r}$
%\item a set of doors it already reached, $D_\mathrm{r}$
%\item a set of spheres other robots have reached $S_\mathrm{o}$
%\item a set of doors other robots have reached $D_\mathrm{o}$
%\end{itemize}
%\State \textbf{Targeting a new door:}
\State $D \gets D\setminus D_\mathrm{r}$, $C \gets C\setminus C_\mathrm{r}$ 
%\State \hspace{2mm} given a set of candidate doors $D$, exclude the doors that were already reached, thus $D \gets D\setminus D_\mathrm{r}$ 
\State \textbf{def} TargetDoor():
\State \hspace{0.7mm}  \textbf{for} each $d \in D$:
\State \hspace{2mm} \textbf{if} $\exists d'\in D_\mathrm{o}
$ $s.t.$ $dist(d,d') < \epsilon_d$: \textbf{ then } $D\gets D\setminus \{d\}$
\State \hspace{1mm} $d_{near}=\arg\min_{d \in D} dist(robot, d); \textbf{return } d_{near}$
%\State \hspace{6mm} exclude this $d$, thus $D \gets D\setminus \{d\}$
%\State \hspace{2mm} choose the closest $d \in D$ as a new target door
%\State \textbf{Targeting a new sphere:}
%\State \hspace{2mm} given a set of candidate spheres $S$, exclude the spheres that were already reached, thus $S \gets S\setminus S_\mathrm{r}$ 
\State \textbf{def} TargetCircle():
\State \hspace{0.7mm} \textbf{for} each $c \in C$:
%\State \hspace{2mm} \textbf{for} each $s \in S$:
\State \hspace{2mm} \textbf{if} $\exists c'\in C_o$ $s.t.$ $dist(c.center,c'.center) < \epsilon_c \cdot c'.r$:
\State \hspace{6mm} \textbf{then} $C \gets C\setminus \{c\}$
\State \hspace{1mm} $c_{near}=\arg\min_{c \in C} dist(robot, c); \textbf{return } c_{near}$
%\State \hspace{2mm} choose the closest $s \in S$ as a new target sphere
%\State \textbf{Inter-Robot Communication:}
\State \textbf{def} Communication($C_r, D_r$):
\State \hspace{2mm} $robot.Publish(C_r, D_r)$
\State \hspace{2mm} $C_o, D_o \gets robot.Subscribe(other\_robots' \hspace{1mm}C_r,D_r)$
\State \hspace{2mm} \textbf{return} $C_o, D_o$
%\State \hspace{2mm} robot publishes spheres $S_\mathrm{r}$ and doors $D_\mathrm{r}$ it reached
%\State \hspace{2mm} robot subscribes to other robots' published $S_\mathrm{r}, D_\mathrm{r}$ and save them to $S_\mathrm{o}$ and $D_\mathrm{o}$
\end{algorithmic}
\end{algorithm}

When targeting a new door, a robot not only excludes doors $D_\mathrm{r}$ reached by itself but also doors $D_\mathrm{o}$ reached by other robots. Any candidate door $d \in D$ that is distanced less than $\epsilon_d$ from any of the doors reached by other robots $D_\mathrm{o}$ is excluded. This is possible because the robots share a common global coordinate frame. Likewise, when targeting a new circle, a robot excludes circles reached by itself and other robots, both $C_\mathrm{r}$ and $C_\mathrm{o}$. The details are explained in TargetDoor() and TargetCircle() methods in Alg.~\ref{alg:target}.

\subsection{Multi-Robot Multi-Room (MRMR) Exploration}
\label{sec:exploration_system}

\begin{algorithm}[t]
\caption{Multi-robot Multi-room (MRMR) exploration. \\ Algorithm is run on each robot at 10Hz. %Lines in \textcolor{teal}{green}/\textcolor{brown}{brown}: the robot \textcolor{teal}{publishes}/\textcolor{brown}{subscribes} to other robots for communication
}\label{alg:MRMR}
\begin{algorithmic}[1]
\Require Initialize $O, C, D, C_\mathrm{r}, D_\mathrm{r}, C_\mathrm{o}, D_\mathrm{o}$ with empty sets $\{\}$
%\Require Initialize $O, C, D, C, C,  spheres $S$ with empty sets $\{\}$, initialize $S_\mathrm{r}, D_\mathrm{r}, S_\mathrm{o}, D_\mathrm{o}$ (as defined in Alg~\ref{alg:target}) with empty sets $\{\}$

%, space partitions $P_i$, viewpoints $V_i$ with $\phi$
\State \textbf{for} each timestep:
%\For{each robot $r_i$}
%\State \textit{$\triangleright$ Mapping}
%\State updates $M_i$ from lidar and odometry sensors
%\State $f_i \gets \text{FrontierExtract}(O_i)$
%\State \textbf{\textit{$\triangleright$ 3D Space Partition}}
%\State \textcolor{blue}{$\vartriangleright$ \textit{geometric analysis of observed map:}}
\State $P^\mathrm{sadd}, (P^\mathrm{max}, \Delta) \gets \textbf{ExtractCues}(O)$ %\Comment{Alg~\ref{alg:extractCues}}
%\State $Q_t \leftarrow $ \text{spheres} $q$ of center $c\in C$ and radius  $\delta \in \Delta$
%\State $S_t \gets $\textbf{$\text{Add-Merge-Spheres}(S_{t-1}, Q_t)$}
\State  $D \gets P^\mathrm{sadd}$;
$C \gets $\textbf{$\text{UpdateCircles}(C,(P^\mathrm{max}, \Delta))$}
%\Comment{Alg~\ref{alg:Spheres}}
%\State \textcolor{blue}{$\vartriangleright$ \textit{door finding:}}
\State \textbf{if not} $robot.is\_targeting$:
\State \hspace{0mm} $d \gets $ \textbf{TargetDoor}(); $robot.is\_targeting \gets True$ 
\State \hspace{1mm}$MoveTo(d)$
%\State\hspace{2mm} target closest door $d \in P^\mathrm{sadd}$ \Comment{Alg~\ref{alg:target}}
%\State \hspace{3mm} \textcolor{teal}{share with other robots: $R$ is targeting $d$}
%\State\hspace{2mm} generate path to $d$ and execute %\textcolor{blue}{\Comment{RRT-Connect}}
%\State \hspace{3mm}
%$\xi \leftarrow \xi + $ path to $d$ \textcolor{blue}{\Comment{RRT-connect plan}}
\State \textbf{if} $robot.reached\_the\_target\_door:$ 
\State \hspace{1mm}$D_\mathrm{r} \leftarrow D_\mathrm{r} \cup \{d\}$
%\textit{\Comment{mark this door as reached}} 
%\State \hspace{2.5mm} \textcolor{teal}{share with other robots: $R$ has reached $d$}
%\State \hspace{3mm}\textcolor{blue}{$\vartriangleright$ \textit{room exploration:}}
\State\hspace{0mm} $c \gets $ \textbf{TargetCircle()}; $MoveTo(c.center)$%target closest sphere $s\in S$ \Comment{Alg~\ref{alg:target}}
%\State \hspace{2.5mm} \textcolor{teal}{share with other robots: $R$ is targeting $s$}
%\textcolor{blue}{\Comment{RRT-Connect}}
%\State \hspace{3mm}
%$\xi \leftarrow \xi + $ path to $s.center$ \textcolor{blue}{\Comment{RRT-connect plan}}
%\State \hspace{2mm} generate and execute path to $s.center$
%\State \hspace{3.5mm}\textbf{if} $room.is\_composed\_of\_multiple\_circles:$
\State \hspace{0mm} \textbf{while} $\exists c'$ $s.t.$ $dist(c'.center, c.center)<\mu(c'.r+c.r)$
\State \hspace{3mm}$c \gets$ \textbf{TargetCircle()}; $MoveTo(c.center)$
%\State \hspace{6mm} target next adjacent sphere $s$ \Comment{Alg~\ref{alg:target}}
%\State \hspace{6mm} generate path to $s.center$ and execute
\State $robot.is\_targeting \gets False$
\State \hspace{0mm}\textbf{for} $\forall c$ $s.t.$ $c.is\_reached$: $C_\mathrm{r} \leftarrow C_\mathrm{r} \cup \{c\}$
%\State \hspace{4mm} $S_\mathrm{r} \leftarrow S_\mathrm{r} \cup \{s\}$ \textit{\Comment{mark spheres as reached}}
\State $C_\mathrm{o}, D_\mathrm{o} \gets$ \textbf{Communication}($C_r, D_r$)
%\State \textcolor{blue}{$\vartriangleright$ \textit{multi-robot communication:}}
%\State update $S_\mathrm{o}, D_\mathrm{o}$ via communication \Comment{Alg~\ref{alg:target}}
%\State \textcolor{teal}{share with other robots: completed-spheres $cS$}
%\State \textcolor{brown}{$oS \leftarrow$ completed/targeted spheres from other robots} 
%\State \textcolor{brown}{$rD \leftarrow$ reached/targeted doors from other robots}
%\State \textcolor{blue}{$\vartriangleright$ \textit{no-target phase:}}
%\State \textbf{if} there is no door or room to target:
%\State \textbf{if} $robot.is\_targeting\_door$  $robot.is\_targeting\_circle$
%\State \hspace{2mm} run baseline \cite{best2022rss} until finding next door
%\State \hspace{3mm}
%$\xi \leftarrow \xi + $ path to random frontiers
\Ensure Path trajectory traveled by robot
\end{algorithmic}
\end{algorithm}

%In this subsection
Here, we explain how the previous algorithms come together and fit into one method, Multi-Robot Multi-Room (MRMR) exploration. The detailed procedure is in Alg.~\ref{alg:MRMR}. Each timestep is a fixed timestep of $0.1s$. At every timestep, the robot extracts geometric cues (saddle points, local maxima, and distances to the walls), and updates the circular decomposition of the space. Then the robot targets a door, moves to the door, and explores a room by targeting a circle that composes the room. When there exists a circle $c'$ that is adjacent to the current circle $c$ (we set $\mu=1.1$, and if distance between $c'.center$ and $c.center$ is smaller than $\mu (c'.r + c.r)$, we regard them adjacent), we set it as a next target circle. This is effective when exploring a larger type of room, composed of multiple adjacent circles. After completing this, the robot marks the doors and circles as reached, and shares this information $D_\mathrm{r}, C_\mathrm{r}$ with other robots to avoid re-targeting of doors and circles that were already explored. This sharing of information through communication enables effective distributed target assignment.

\begin{figure}[!t]
    \centering
\includegraphics[width=0.85\linewidth, height=0.35\linewidth]{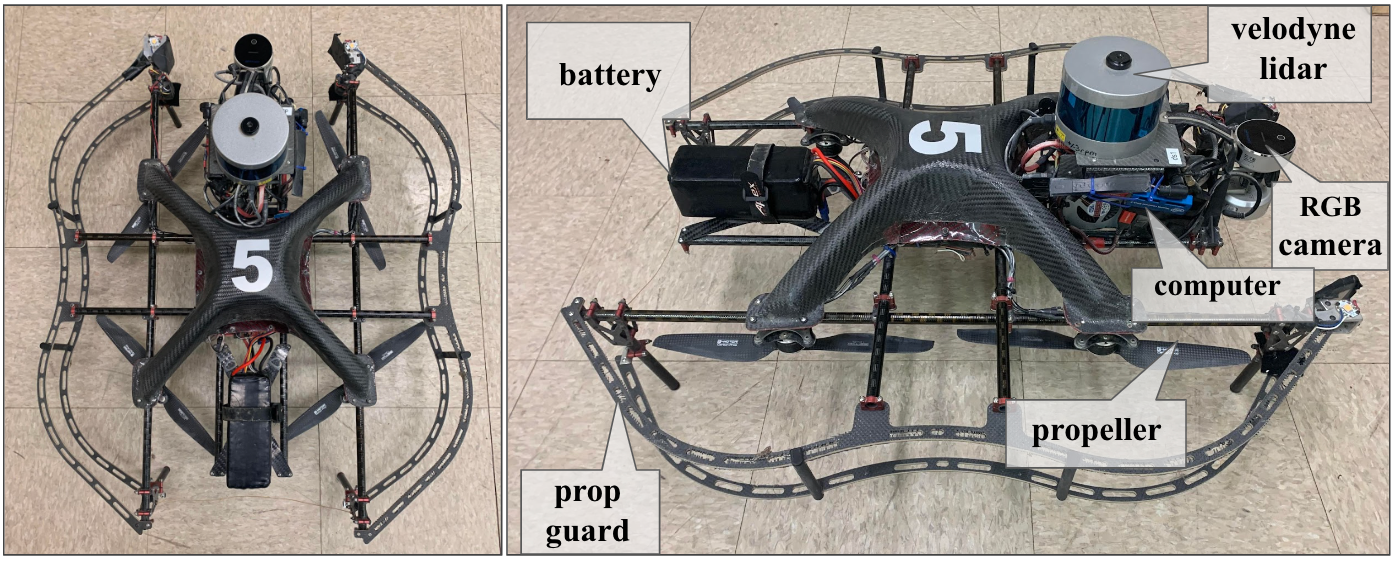}
    \caption{Aerial robot system hardware. Left: top-view, Right: side-view. Custom-built quadrotor robot equipped with sensors and computer onboard.}
    \label{drone}
\end{figure}

\section{Experiments \& Results}
In this section, we explain the experimental setup and display results to evaluate our algorithm, MRMR. We first elaborate on the experimental setup and discuss results for simulation environments. Then we discuss the experimental setup and results for real-robot experiments. For comparison, we chose the previous frontier-based exploration \cite{best2022rss} as a baseline, both in simulation and real-robot experiments. 

\subsection{Simulation Experiments}

\begin{table}[]
    \centering
        \caption{The details of the simulation test environments. }
    \newlength{\mapspace}
    \setlength{\mapspace}{2pt}
    \begin{tabularx}{\linewidth}{cXccc}%{|p{1cm}|p{3cm}|p{1cm}|p{1.7cm}|}
       \toprule
       Name & \hspace{0.8cm}Images  & Rooms & Volume($m^3$) & Voxels\\
       \midrule
       Env1 & \raisebox{-0.5\height}{%
       \includegraphics[width=2.5cm]{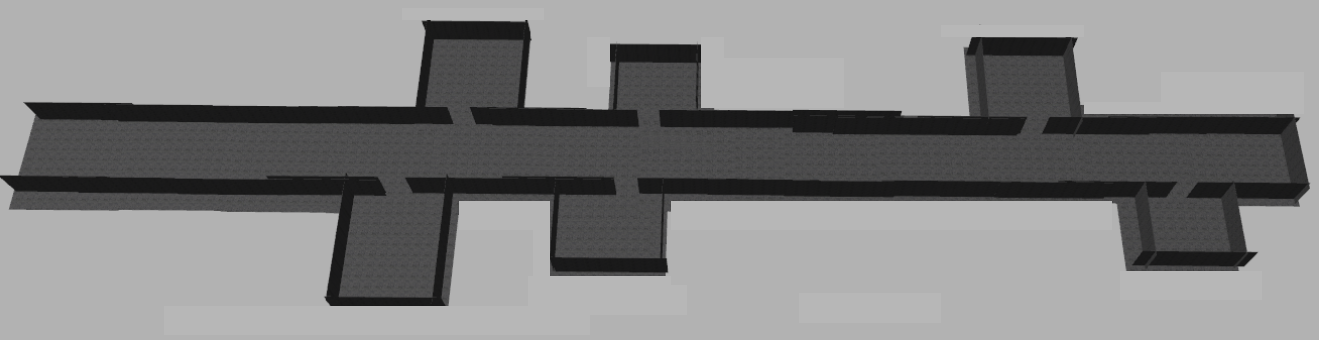}}  & 6 & 462.6 & 9587 \\
       %\addlinespace[\mapspace]
       Env2 & \raisebox{-0.5\height}{%
       \includegraphics[width=2.5cm,height=0.8cm]{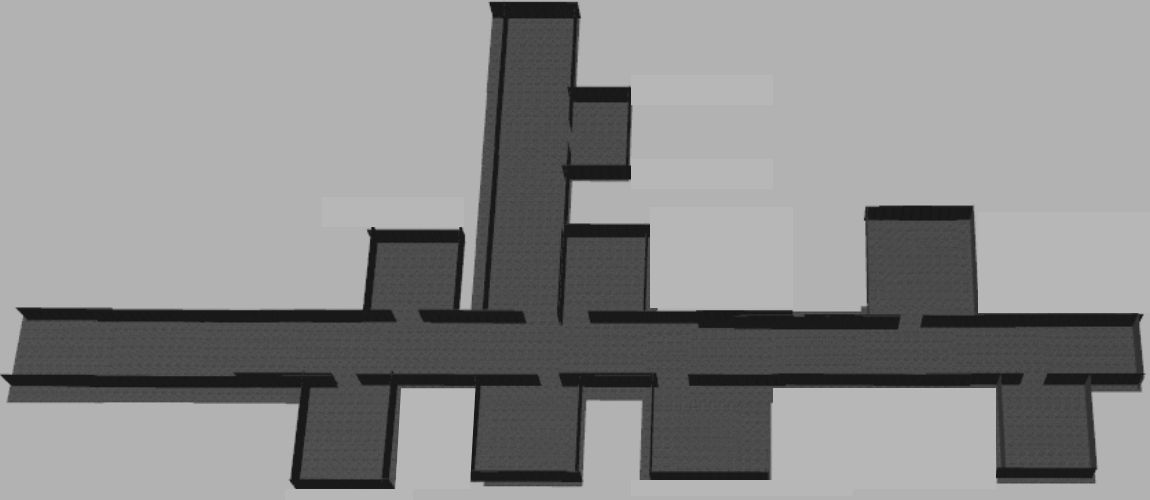}} & 8 & 776.7 & 21313 \\
       %\addlinespace[\mapspace]
       Env3 & \raisebox{-0.5\height}{%
      \includegraphics[width=2.5cm,height=0.8cm]{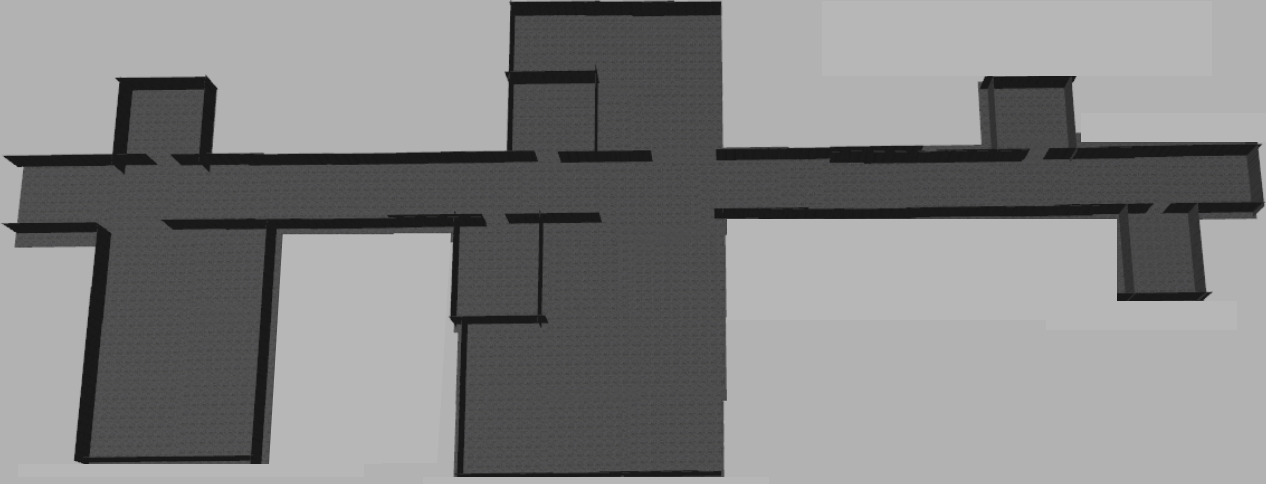}}  & 5 & 402.9 & 11869 \\
       %\addlinespace[\mapspace]
       Env4 & \raisebox{-0.5\height}{%
      \includegraphics[width=2.5cm,height=0.8cm]{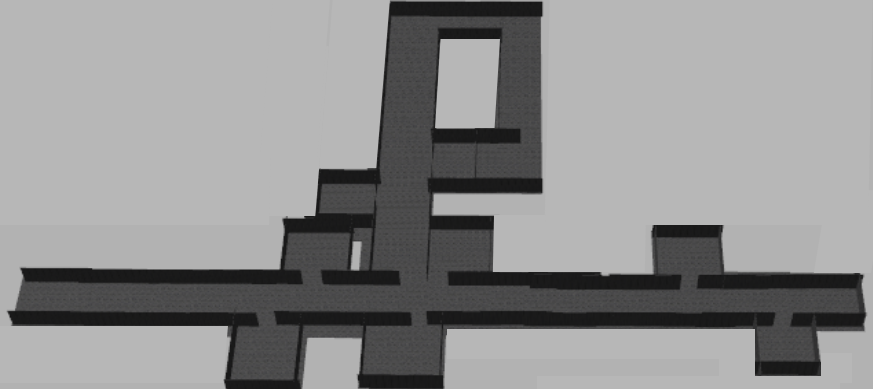}}  & 9 & 729.1 & 20105 \\
       %\addlinespace[\mapspace]
       Env5 & \raisebox{-0.5\height}{%
      \includegraphics[width=2.5cm,height=0.9cm]{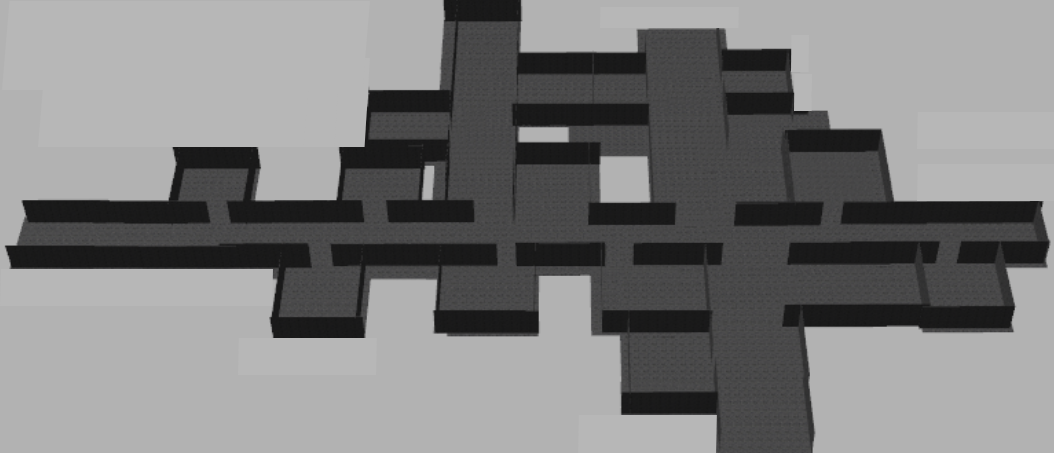}}  & 12 & 1179.1 & 33371 \\
       %\addlinespace[\mapspace]
      Env6 & \raisebox{-0.5\height}{%
      \includegraphics[width=2.5cm,height=0.85cm]{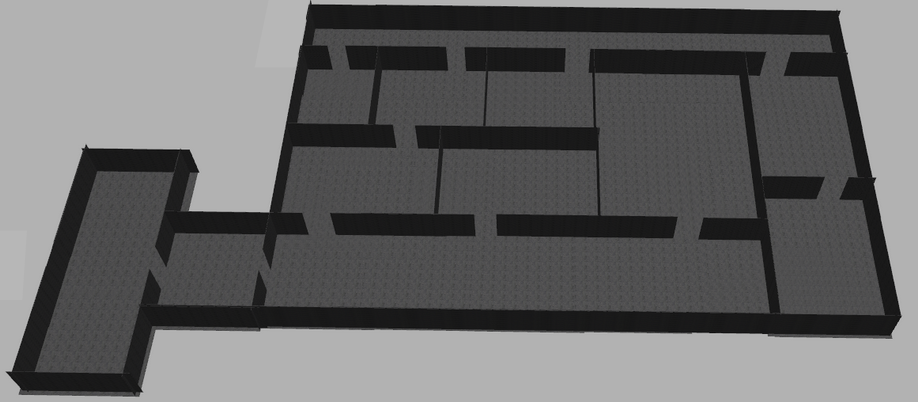}}  & 9 & 1226.9 & 38807 \\
      %\addlinespace[\mapspace]
       Env7 & \raisebox{-0.5\height}{%
      \includegraphics[width=2.5cm,height=0.95cm]{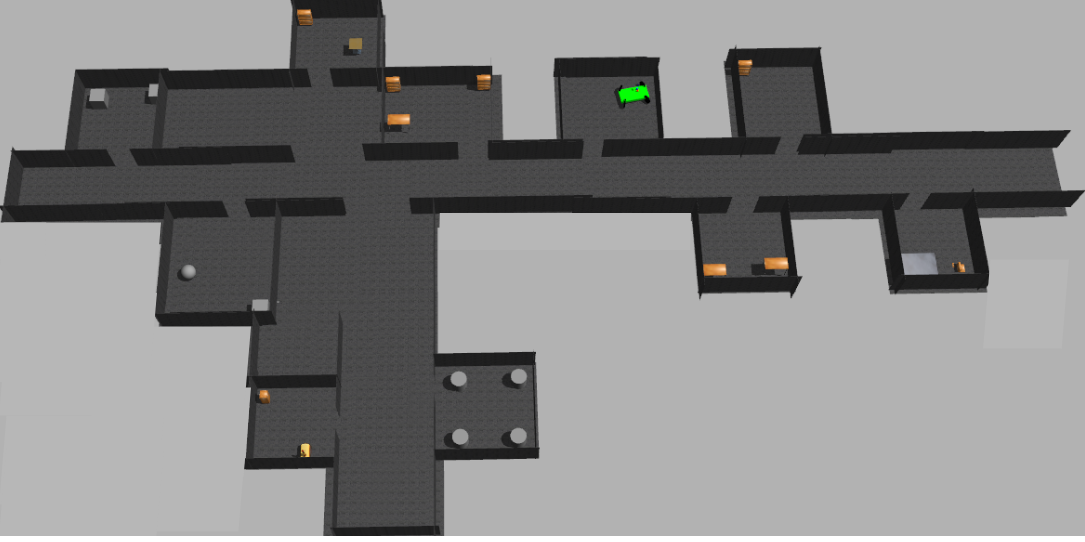}}  & 10 & 932.4 & 23734 \\
      %\addlinespace[\mapspace]
       Env8 & \raisebox{-0.5\height}{%
      \includegraphics[width=2.5cm,height=1.0cm]{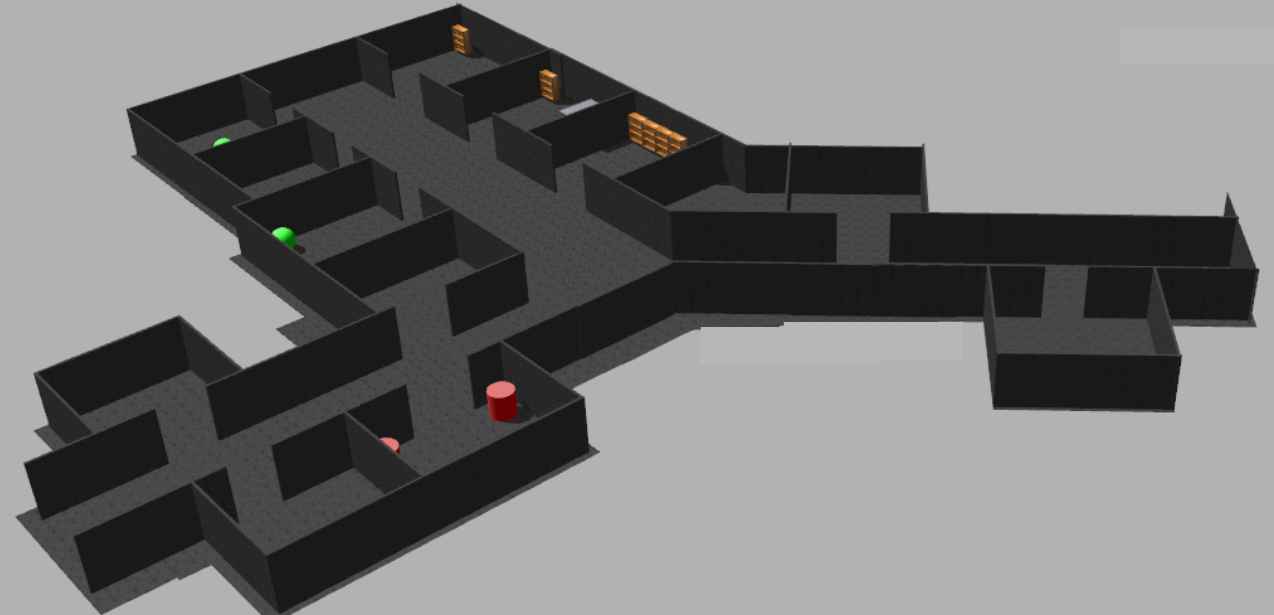}}  & 12 & 1024.9 & 31082 \\
       \bottomrule
    \end{tabularx}
    \label{tab:sim-environments}
\end{table}

\subsubsection{Experimental Setup}
\begin{figure*}[!t]
    \centering
\includegraphics[width=1.0\linewidth, height=0.32\linewidth]{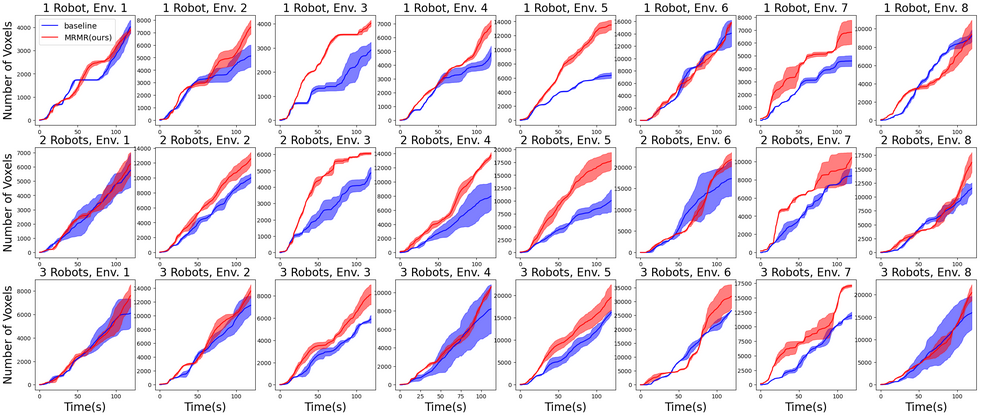}
    \caption{The comparison of baseline and our method (MRMR). The first, second, third row is 1-robot, 2-robots, and 3-robots case respectively, and each column represents each different environment. Results show that our MRMR method observes more voxels in rooms than the baseline.}
    \label{sim-graphs}
\end{figure*}
% We divide this section into simulation experiments and real-world experiments. 
We used the SubT UAV code released by \cite{best2022rss}, developed to simulate real subterranean environments. We designed eight indoor simulation environments with different configurations; Table~\ref{tab:sim-environments} displays the details of each environment. %Each environment is composed of multiple rooms, with different lengths of corridors, different sizes of halls, and different complexities of floorplan. 
First five environments (Env1 to Env5) feature simple, idealized settings with doors, rooms and corridors. Env6 features larger sizes of rooms. Env7 includes objects (e.g. table, bookshelf, cabinet) in each room. Lastly, Env8 includes a scenario where the wall-axes are not aligned with x,y-axis of the global coordinate frame.  

We tested with different numbers of robots ($n=1,2,3$) and measured the number of voxels collectively observed by the robots' camera sensors over 2 minutes. Each graph curve is an average of three independent runs. We also measured percentages of observed voxels (out of total voxels in rooms) and number of rooms (out of total number of rooms) by each method, and displayed the results in a separate table.

\begin{figure}[!t]
    \centering
\includegraphics[width=0.9\linewidth]{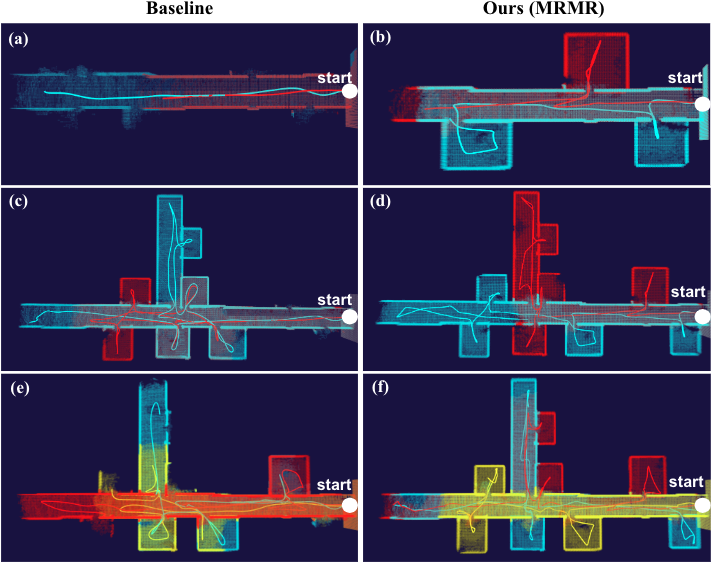}
    \caption{Visualization of baseline's and our method's behaviors. (a) Using the baseline, the robots travel the corridor first by moving straight, while (b) using our method, robots quickly turn directions to find doors and enter the rooms. (c),(e): multiple robots often enter the same rooms redundantly. (d),(f): each room is uniquely visited by each robot using our method.}
    \label{sim-visualize}
\end{figure}

\subsubsection{Results}

\begin{table}[]
    \centering
        \caption{ The summary of simulation experiments}
        %\gb{needs to fit in column -- either play with spacing or re-introduce the abbreviations. in either case, add more explaination to this caption}
    \begin{tabular}
{p{1.35cm}p{0.72cm}p{0.72cm}p{0.72cm}p{0.72cm}p{0.72cm}p{0.72cm}}
        \toprule
        \textbf{Method} & \multicolumn{2}{c}{\textbf{1 Robot}} & \multicolumn{2}{c}{\textbf{2 Robots}}  & \multicolumn{2}{c}{\textbf{3 Robots}} \\
        & Vxl. & Rm. & Vxl. & Rm. & Vxl. & Rm. \\
       \midrule
       Baseline & 27.15\% & 19.17\% & 39.08\% & 41.14\% & 52.88\% & 58.76\% \\
       MRMR & \textbf{35.57\%} & \textbf{34.58\%} & \textbf{55.43\%} & \textbf{51.49\%} & \textbf{67.4\%} & \textbf{66.68\%} \\
       \addlinespace[2pt]
       \emph{Improvement} & 31.01\% & 80.38\% & 41.84\% & 25.16\% & 27.45\% & 13.48\% \\
       \bottomrule
    \end{tabular}
    \label{tab:sim-summary}
\end{table}

The results of the simulation experiments are displayed in Fig.~\ref{sim-graphs}. The plots in the first row of the figure are single-robot case, and the plots in the second and third row are multi-robot cases (2-robots and 3-robots). These plots demonstrate that our Method (MRMR) observes more voxels in the rooms than frontier-based baseline, across different environments and different number of robots. In average, MRMR observes $31.01\%, 41.84\%, 27.45\%$ more room grid voxels than baseline in single-robot, two-robots, and three-robots cases. In the Table~\ref{tab:sim-summary}, we reported how MRMR observes more voxels (out of total room voxels; denoted Vxl.) and explore more rooms (out of total number of rooms; denoted Rm.) than baseline; MRMR outperforms the baseline both in terms of voxels and number of rooms. MRMR shows larger outperformance compared to baseline in the case of two-robots than three-robots. While outperformance is still significant, adding more robots to the same space results in necessary overlap of their sensor coverage.

More qualitative explanations of difference between the baseline and MRMR are following: as shown in Fig.~\ref{sim-visualize}(a), using the baseline, robots first choose to travel corridors fast by moving straight to increase the coverage area without turning to the doors or rooms. In contrary, as in Fig.~\ref{sim-visualize}(b), the robots quickly turn directions to reach the doors and enter rooms, rather than traveling along the corridor. Using the baseline (Fig.~\ref{sim-visualize}(c),(e)), multiple robots often enter the same rooms redundantly and sometimes miss exploring rooms; in contrary our method (Fig.~\ref{sim-visualize}(d),(f)) enables robots to uniquely visit each room without redundancy. 

We briefly report the measured runtime of new modules. The average runtimes of distance transform, geometric cue extraction, and circular decomposition are $480.5\mu s$, $1.962 ms$, and $35.8\mu s$. We also report the accuracy of door detection using saddle points extraction: The average precision and recall over all environments are $0.94$ and $0.97$.

\begin{figure}[!t]
    \centering
\includegraphics[width=0.9\linewidth]{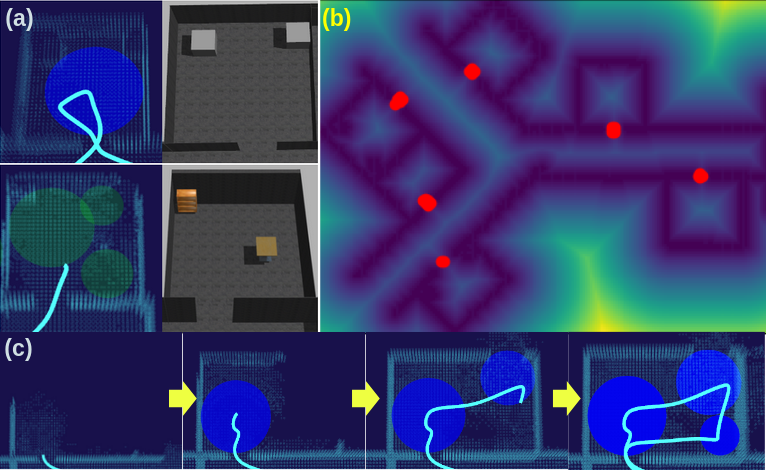}
    \caption{(a) When objects are in rooms, the robot generates circles that adjoin the objects. (b) Our saddle point detection is robust to the case where door axis is not aligned with global coordinate frame. (c) Robot generates multiple circles while exploring a large room.}
    \label{challenging-environments}
\end{figure}

Lastly, we discuss the behaviors of robots in challenging environments. When objects are in rooms, (Fig.~\ref{challenging-environments}(a)), robots generate circles that are tangential to the objects, as these objects serve as occupied cells when generating 2D distance transform map. When wall axes are not aligned with global frame (Fig~\ref{challenging-environments}(b)), our saddle point detection is still robust and accurate at detecting doors. In the setting with a larger room (Fig.~\ref{challenging-environments}(c)), the robot generates multiple circles and reaches the centers of the circles one by one to cover the room.

\subsection{Real-Robot Experiments}

\subsubsection{Experimental Setup}

For real-world experiments, we use custom quadrotors as shown in  Fig.~\ref{drone}. Each drone is 68cm wide, 81cm long, weighs 5.2kg, and is equipped with a Velodyne VLP-16 Lidar, Realsense RGBD sensors, Rajant DX2, Intel NUC computer with Intel Core i7-8550U CPU. No additional data filtering or clean-ups are required compared to simulation. We run experiments in an abandoned hospital in Pittsburgh, PA. The main results compare our method to the baseline with two to three robots; drones fly sequentially as the mechanism for inter-robot collision avoidance for real robots is limited. Additionally, we report results of a trial with three robots flying simultaneously with our method only. We report the number of voxels (Vxl.) and number of rooms (Rm.) collectively observed by the robots for both experiments.

\subsubsection{Results}

\begin{table}[]
    \centering
    %\gb{needs to fit in column -- either play with spacing or re-introduce the abbreviations. in either case, add more explaination to this caption}
        \caption{The summary of real-robot experiments over three trials}
    \begin{tabularx}{\linewidth}{lcccccccc}
    \toprule
       & \multicolumn{4}{c}{\textbf{2 Robots}} & 
       \multicolumn{4}{c}{\textbf{3 Robots}} \\
       \cmidrule(lr){2-5}\cmidrule(lr){6-9}
       \# & \multicolumn{2}{c} {Baseline} & \multicolumn{2}{c} {MRMR} & 
       \multicolumn{2}{c} {Baseline} & 
       \multicolumn{2}{c} {MRMR}
 \\
   & Vxl. & Rm. & Vxl. & Rm. & Vxl. & Rm. & Vxl. & Rm. \\ 
 \midrule
       1. & 6204 & 3 & \textbf{10285} & \textbf{5} & 11584 & 5 & \textbf{15732} & \textbf{8} \\
      2. & \textbf{10721} & 5 & 9428 & 5 & 14336 & 7 & \textbf{17320} & \textbf{9} \\
      3. & 8437 & 4 & \textbf{11724} & \textbf{6} & 14561 & 7 & \textbf{15976} & \textbf{8} \\
       \bottomrule
    \end{tabularx}
    \label{tab:real-summary}
\end{table}

\begin{figure}[!t]
    \centering
\includegraphics[width=1.0\linewidth]{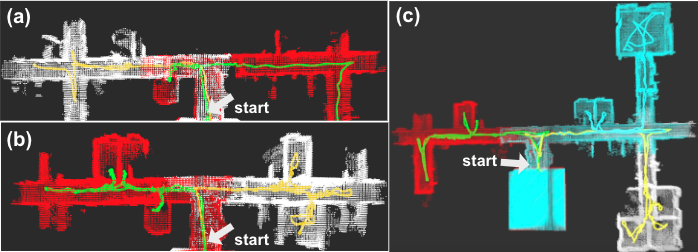}
    \caption{Visualization of covered areas and traveled trajectories in real-robot experiments: (a) baseline (two-robots) (b) MRMR (two-robots). MRMR observes more rooms than baseline. (c) MRMR (three-robots)}
    \label{real-robot-visualize}
\end{figure}

The results are displayed in the Table~\ref{tab:real-summary}. In the comparative tests, both in two-robots and three-robots cases, the robots generally observe more voxels and explore more rooms using MRMR. The exception is trial 2 of the two-robot case where the baseline outperforms MRMR but by an insignificant amount.
On average, MRMR observes $30.66\%, 22.09\%$ more voxels in rooms than the baseline, in two-robot and three-robot cases respectively. Fig.~\ref{real-robot-visualize}(a),(b) are the results of running both methods with two robots. Finally, in the simultaneous execution experiment (Fig.~\ref{real-robot-visualize}(c)), the robots successfully explored 8 rooms using MRMR, observing $16793$ voxels in the rooms, with three drones.

% \section{RESULTS}

% Text heads organize the topics on a relational, hierarchical basis. For example, the paper title is the primary text head because all subsequent material relates and elaborates on this one topic. If there are two or more sub-topics, the next level head (uppercase Roman numerals) should be used and, conversely, if there are not at least two sub-topics, then no subheads should be introduced. Styles named ÒHeading 1Ó, ÒHeading 2Ó, ÒHeading 3Ó, and ÒHeading 4Ó are prescribed.
%\begin{figure}[!t]
%    \centering
%\includegraphics[width=1.0\linewidth]%{Figures/temp-single-robot-graphs.png}
%    \caption{single-robot}
%    \label{single-robot graph goes here}
%\end{figure}
%\begin{figure*}[!t]
%    \centering
%\includegraphics[width=1.0\textwidth]%{Figures/two-three-robot-graphs.png}
%    \caption{Two robots and three robots case experiment results go here}
%    \label{fig:two-three-robots}
%\end{figure*}

% \begin{figure}[!t]
%     \centering
% \includegraphics[width=0.9\linewidth]{Figures/real-three-drone-test-hawkins.png}
%     \caption{Three robots are \textit{simultaneously} exploring rooms in real-drone experiments; the trajectory and area covered by each robot is visualized with different color.}
%     \label{real-visualize}
% \end{figure}

\section{Conclusion \& Future Work}
We proposed a multi-robot multi-room autonomous exploration pipeline (MRMR) that methodically explores rooms in a building and coordinates behaviors of robots. To this end, we presented a geometric cue extraction method that detects locations of doors and rooms from point clouds, and circular decomposition of spaces for target assignment. We validated performance of our method in simulated and real experiments.
Some limitations are that the approach is only applicable to single story buildings due to flattening of 3D voxels into 2D distance transform, and that the collision avoidance among aerial robots is implicit, rather than explicit. For future work, we plan to extend our approach to multi-floor buildings and to consider exploration with heterogeneous multi-robot systems and to improve such systems with advanced vision and learning modules.

% The balance command serves this purpose
%\addtolength{\textheight}{-12cm}   % This command serves to balance the column lengths
                                  % on the last page of the document manually. It shortens
                                  % the textheight of the last page by a suitable amount.
                                  % This command does not take effect until the next page
                                  % so it should come on the page before the last. Make
                                  % sure that you do not shorten the textheight too much.
{
    \bibliographystyle{IEEEtranN}
    \scriptsize
    \balance
    \bibliography{IEEEabrv, IEEEexample}

% Generated by IEEEtranN.bst, version: 1.14 (2015/08/26)
\begin{thebibliography}{37}
\providecommand{\natexlab}[1]{#1}
\providecommand{\url}[1]{#1}
\csname url@samestyle\endcsname
\providecommand{\newblock}{\relax}
\providecommand{\bibinfo}[2]{#2}
\providecommand{\BIBentrySTDinterwordspacing}{\spaceskip=0pt\relax}
\providecommand{\BIBentryALTinterwordstretchfactor}{4}
\providecommand{\BIBentryALTinterwordspacing}{\spaceskip=\fontdimen2\font plus
\BIBentryALTinterwordstretchfactor\fontdimen3\font minus
  \fontdimen4\font\relax}
\providecommand{\BIBforeignlanguage}[2]{{%
\expandafter\ifx\csname l@#1\endcsname\relax
\typeout{** WARNING: IEEEtranN.bst: No hyphenation pattern has been}%
\typeout{** loaded for the language `#1'. Using the pattern for}%
\typeout{** the default language instead.}%
\else
\language=\csname l@#1\endcsname
\fi
#2}}
\providecommand{\BIBdecl}{\relax}
\BIBdecl

\bibitem[Burgard et~al.(2005)Burgard, Moors, Stachniss, and
  Schneider]{Burgard2005}
W.~Burgard, M.~Moors, C.~Stachniss, and F.~E. Schneider, ``Coordinated
  {M}ulti-{R}obot {E}xploration,'' \emph{{IEEE} Trans. Robotics}, vol.~21,
  no.~3, pp. 376--386, 2005.

\bibitem[Braga et~al.(2017)Braga, Aguiar, and de~Sousa]{braga2017}
J.~Braga, A.~P. Aguiar, and J.~B. de~Sousa, ``Coordinated {M}ulti-{UAV}
  {E}xploration {S}trategy for {L}arge {A}reas with {C}ommunication
  {C}onstrains,'' in \emph{ROBOT 2017: Third Iberian Robotics Conference:
  Volume 1}, 2017, pp. 149--160.

\bibitem[Queralta et~al.(2020)]{search-and-rescue}
J.~P. Queralta \emph{et~al.}, ``Collaborative {M}ulti-{R}obot {S}earch and
  {R}escue: Planning, {C}oordination, {P}erception, and {A}ctive {V}ision,''
  \emph{IEEE Access}, 2020.

\bibitem[Ristic et~al.(2017)]{hazardous}
B.~Ristic \emph{et~al.}, ``Autonomous {M}ulti-{R}obot {S}earch for a
  {H}azardous {S}ource in a {T}urbulent {E}nvironment,'' \emph{Sensors},
  vol.~17, no.~4, p. 918, 2017.

\bibitem[Schuster et~al.(2019)]{schuster2019towards}
M.~J. Schuster \emph{et~al.}, ``Towards {A}utonomous {P}lanetary
  {E}xploration,'' \emph{Journal of Intelligent \& Robotic Systems}, vol.~93,
  pp. 461--494, 2019.

\bibitem[Kulkarni et~al.(2022)]{kulkarni2022subt}
M.~Kulkarni \emph{et~al.}, ``Autonomous {T}eamed {E}xploration of
  {S}ubterranean {E}nvironments using {L}egged and {A}erial {R}obots,'' in
  \emph{2022 International Conference on Robotics and Automation (ICRA)}, 2022,
  pp. 3306--3313.

\bibitem[Scherer et~al.(2022)]{Scherer:2022}
S.~Scherer \emph{et~al.}, ``Resilient and {M}odular {S}ubterranean
  {E}xploration with a {T}eam of {R}oving and {F}lying {R}obots,'' \emph{Field
  Robotics Journal}, pp. 678--734, May 2022.

\bibitem[Agha et~al.(2021)Agha, Otsu, Morrell, Fan, Thakker,
  et~al.]{agha2021nebula}
A.~Agha, K.~Otsu, B.~Morrell, D.~D. Fan, R.~Thakker \emph{et~al.}, ``Nebula:
  Quest for {R}obotic {A}utonomy in {C}hallenging {E}nvironments; {TEAM}
  {CoSTAR} at the {DARPA} {S}ubterranean {C}hallenge,'' \emph{arXiv preprint
  arXiv:2103.11470}, 2021.

\bibitem[Yamauchi(1998)]{yamauchi1998frontier}
B.~Yamauchi, ``Frontier-{B}ased {E}xploration using {M}ultiple {R}obots,'' in
  \emph{The 2nd International Conference on Autonomous Agents}, 1998, pp.
  47--53.

\bibitem[Best et~al.(2022)Best, Garg, Keller, Hollinger, and
  Scherer]{best2022rss}
G.~Best, R.~Garg, J.~Keller, G.~A. Hollinger, and S.~Scherer, ``Resilient
  {M}ulti-{S}ensor {E}xploration of {M}ultifarious {E}nvironments with a {T}eam
  of {A}erial {R}obots,'' in \emph{Proceedings of Robotics: Science and
  Systems}, 2022.

\bibitem[Hollinger and Sukhatme(2013)]{sample-based-motion-planning}
G.~A. Hollinger and G.~S. Sukhatme, ``Sampling-based {M}otion {P}lanning for
  {R}obotic {I}nformation {G}athering,'' in \emph{Robotics: Science and
  Systems}, 2013.

\bibitem[Charrow(2015)]{charrow2015information}
B.~Charrow, \emph{Information-{T}heoretic {A}ctive {P}erception for
  {M}ulti-{R}obot {T}eams}.\hskip 1em plus 0.5em minus 0.4em\relax University
  of Pennsylvania, 2015.

\bibitem[Corah and Michael(2019)]{submodular-micah2019}
M.~Corah and N.~Michael, ``Distributed {M}atroid-{C}onstrained {S}ubmodular
  {M}aximization for {M}ulti-{R}obot {E}xploration: Theory and {P}ractice,''
  \emph{Autonomous Robots}, vol.~43, pp. 485--501, 2019.

\bibitem[Dang et~al.(2020)Dang, Tranzatto, Khattak, Mascarich, Alexis, and
  Hutter]{dang2020graph}
T.~Dang, M.~Tranzatto, S.~Khattak, F.~Mascarich, K.~Alexis, and M.~Hutter,
  ``Graph-based {S}ubterranean {E}xploration {P}ath {P}lanning using {A}erial
  and {L}egged {R}obots,'' \emph{Journal of Field Robotics}, vol.~37, no.~8,
  pp. 1363--1388, 2020.

\bibitem[Sung and Tokekar(2019)]{plume2019}
Y.~Sung and P.~Tokekar, ``A {C}ompetitive {A}lgorithm for {O}nline
  {M}ulti-{R}obot {E}xploration of a {T}ranslating {P}lume,'' in
  \emph{International Conference on Robotics and Automation, {ICRA}}, 2019, pp.
  3391--3397.

\bibitem[Yan et~al.(2022)Yan, Lin, Ren, Zhao, Yu, Cao, Yin, Zhang, and
  Scherer]{MUI-TARE}
J.~Yan, X.~Lin, Z.~Ren, S.~Zhao, J.~Yu, C.~Cao, P.~Yin, J.~Zhang, and S.~A.
  Scherer, ``{MUI-TARE:} {M}ulti-{A}gent {C}ooperative {E}xploration with
  {U}nknown {I}nitial {P}osition,'' \emph{CoRR}, vol. abs/2209.10775, 2022.

\bibitem[Luna and Bekris(2011)]{luna2011efficient}
R.~Luna and K.~E. Bekris, ``Efficient and {C}omplete {C}entralized
  {M}ulti-{R}obot {P}ath {P}lanning,'' in \emph{IEEE/RSJ International
  Conference on Intelligent Robots and Systems}, 2011, pp. 3268--3275.

\bibitem[Gul et~al.(2022)Gul, Mir, Mir, Mir, Islaam, Abualigah, and
  Forestiero]{gul2022centralized}
F.~Gul, A.~Mir, I.~Mir, S.~Mir, T.~U. Islaam, L.~Abualigah, and A.~Forestiero,
  ``A {C}entralized {S}trategy for {M}ulti-{A}gent {E}xploration,'' \emph{IEEE
  Access}, vol.~10, pp. 126\,871--126\,884, 2022.

\bibitem[Li et~al.(2020)Li, Gama, Ribeiro, and Prorok]{li2020graph}
Q.~Li, F.~Gama, A.~Ribeiro, and A.~Prorok, ``Graph {N}eural {N}etworks for
  {D}ecentralized {M}ulti-{R}obot {P}ath {P}lanning,'' in \emph{IEEE/RSJ
  International Conference on Intelligent Robots and Systems}, 2020, pp.
  11\,785--11\,792.

\bibitem[Zhai et~al.(2021)Zhai, Ding, Liu, Jia, Zhao, and
  Luo]{zhai2021decentralized}
Y.~Zhai, B.~Ding, X.~Liu, H.~Jia, Y.~Zhao, and J.~Luo, ``Decentralized
  {M}ulti-{R}obot {C}ollision {A}voidance in {C}omplex {S}cenarios with
  {S}elective {C}ommunication,'' \emph{IEEE Robotics and Automation Letters},
  vol.~6, no.~4, pp. 8379--8386, 2021.

\bibitem[Wu et~al.(2007)]{voronoi-based-space-partitioning}
L.~Wu \emph{et~al.}, ``Voronoi-based {S}pace {P}artitioning for {C}oordinated
  {M}ulti-{R}obot {E}xploration,'' \emph{Journal of Physical Agents}, 2007.

\bibitem[Solanas and Garcia(2004)]{unsup-clustering-space}
A.~Solanas and M.~Garcia, ``Coordinated {M}ulti-{R}obot {E}xploration through
  {U}nsupervised {C}lustering of {U}nknown {S}pace,'' in \emph{IEEE/RSJ
  International Conference on Intelligent Robots and Systems}, 2004.

\bibitem[Hu et~al.(2020)Hu, Niu, Carrasco, Lennox, and Arvin]{Voronoi-deep-rl}
J.~Hu, H.~Niu, J.~Carrasco, B.~Lennox, and F.~Arvin, ``Voronoi-based
  {M}ulti-{R}obot {A}utonomous {E}xploration in {U}nknown {E}nvironments via
  {D}eep {R}einforcement {L}earning,'' \emph{{IEEE} Trans. Veh. Technol.},
  vol.~69, no.~12, pp. 14\,413--14\,423, 2020.

\bibitem[Gao et~al.(2019)Gao, Wu, Gao, and Shen]{Gao19}
F.~Gao, W.~Wu, W.~Gao, and S.~Shen, ``Flying on {P}oint {C}louds: {O}nline
  {T}rajectory {G}eneration and {A}utonomous {N}avigation for {Q}uadrotors in
  {C}luttered {E}nvironments,'' \emph{J. Field Robotics}, vol.~36, no.~4, pp.
  710--733, 2019.

\bibitem[Ren et~al.(2022)Ren, Zhu, Liu, Wang, Lin, Gao, and
  Zhang]{bubble-planner}
Y.~Ren, F.~Zhu, W.~Liu, Z.~Wang, Y.~Lin, F.~Gao, and F.~Zhang, ``Bubble
  {P}lanner: {P}lanning {H}igh-speed {S}mooth {Q}uadrotor {T}rajectories using
  {R}eceding {C}orridors,'' in \emph{{IEEE/RSJ} International Conference on
  Intelligent Robots and Systems, {IROS} 2022, Kyoto, Japan, October 23-27,
  2022}.\hskip 1em plus 0.5em minus 0.4em\relax {IEEE}, 2022, pp. 6332--6339.

\bibitem[Musil et~al.(2022)Musil, Petrl{\'{\i}}k, and Saska]{SphereMap}
T.~Musil, M.~Petrl{\'{\i}}k, and M.~Saska, ``Spheremap: {D}ynamic
  {M}ulti-{L}ayer {G}raph {S}tructure for {R}apid {S}afety-{A}ware {UAV}
  {P}lanning,'' \emph{{IEEE} Robotics Autom. Lett.}, vol.~7, no.~4, pp.
  11\,007--11\,014, 2022.

\bibitem[Thrun(1998)]{thrun1998}
S.~Thrun, ``Learning {M}etric-{T}opological {M}aps for {I}ndoor {M}obile
  {R}obot {N}avigation,'' \emph{Artificial Intelligence}, vol.~99, no.~1, pp.
  21--71, 1998.

\bibitem[Wurm et~al.(2008)Wurm, Stachniss, and Burgard]{wurm2008coordinated}
K.~M. Wurm, C.~Stachniss, and W.~Burgard, ``Coordinated {M}ulti-{R}obot
  {E}xploration using a {S}egmentation of the {E}nvironment,'' in
  \emph{IEEE/RSJ International Conference on Intelligent Robots and Systems},
  2008, pp. 1160--1165.

\bibitem[Brunskill et~al.(2007)Brunskill, Kollar, and
  Roy]{brunskill2007topological}
E.~Brunskill, T.~Kollar, and N.~Roy, ``Topological {M}apping using {S}pectral
  {C}lustering and {C}lassification,'' in \emph{IEEE/RSJ International
  Conference on Intelligent Robots and Systems}, 2007, pp. 3491--3496.

\bibitem[Oleynikova et~al.(2018)]{sparse-3d-topologicalg-graphs}
H.~Oleynikova \emph{et~al.}, ``Sparse 3d topological graphs for micro-aerial
  vehicle planning,'' in \emph{2018 {IEEE/RSJ} International Conference on
  Intelligent Robots and Systems, {IROS} 2018, Madrid, Spain, October 1-5,
  2018}.\hskip 1em plus 0.5em minus 0.4em\relax {IEEE}, 2018, pp. 1--9.

\bibitem[Sj{\"o}{\"o}(2012)]{sjoo2012semantic}
K.~Sj{\"o}{\"o}, ``Semantic {M}ap {S}egmentation using {F}unction-based
  {E}nergy {M}aximization,'' in \emph{IEEE International Conference on Robotics
  and Automation}, 2012.

\bibitem[Bavle et~al.(2023)Bavle, S{\'{a}}nchez{-}L{\'{o}}pez, Shaheer, Civera,
  and Voos]{S-graphs+}
H.~Bavle, J.~L. S{\'{a}}nchez{-}L{\'{o}}pez, M.~Shaheer, J.~Civera, and
  H.~Voos, ``S-graphs+: Real-time localization and mapping leveraging
  hierarchical representations,'' \emph{{IEEE} Robotics Autom. Lett.}, vol.~8,
  no.~8, pp. 4927--4934, 2023.

\bibitem[Rosinol et~al.(2020)Rosinol, Gupta, Abate, Shi, and
  Carlone]{3d-dynamic-scene-graphs}
A.~Rosinol, A.~Gupta, M.~Abate, J.~Shi, and L.~Carlone, ``{3D} {D}ynamic
  {S}cene {G}raphs: {A}ctionable {S}patial {P}erception with {P}laces,
  {O}bjects, and {H}umans,'' in \emph{Robotics: Science and Systems XVI}, 2020.

\bibitem[Hughes et~al.(2022)Hughes, Chang, and Carlone]{Hydra}
N.~Hughes, Y.~Chang, and L.~Carlone, ``Hydra: {A} real-time spatial perception
  system for 3d scene graph construction and optimization,'' in \emph{Robotics:
  Science and Systems XVIII, New York City, NY, USA, June 27 - July 1, 2022},
  2022.

\bibitem[Museth(2013)]{open-vdb}
K.~Museth, ``{VDB}: High-{R}esolution {S}parse {V}olumes with {D}ynamic
  {T}opology,'' \emph{ACM transactions on graphics (TOG)}, vol.~32, no.~3, pp.
  1--22, 2013.

\bibitem[Zhao et~al.(2021)]{super-odometry}
S.~Zhao \emph{et~al.}, ``Super {O}dometry: {IMU}-centric
  {LiDAR}-{V}isual-{I}nertial {E}stimator for {C}hallenging {E}nvironments,''
  in \emph{IEEE/RSJ International Conference on Intelligent Robots and
  Systems}, 2021, pp. 8729--8736.

\bibitem[Kuffner and LaValle(2000)]{RRT-connect}
J.~J. Kuffner and S.~M. LaValle, ``{RRT}-{C}onnect: {A}n {E}fficient {A}pproach
  to {S}ingle-{Q}uery {P}ath {P}lanning,'' in \emph{Proceedings of IEEE
  International Conference on Robotics and Automation.}, vol.~2, 2000, pp.
  995--1001.

\end{thebibliography}
}

\end{document}